\documentclass{opt2024} 

\usepackage{adjustbox}
\usepackage{mathrsfs}
\usepackage{mdframed}
\usepackage{float}
\usepackage{multirow}
\usepackage{changepage}      

\usepackage{graphicx} 
\usepackage{parskip}  
\usepackage{setspace} 
\usepackage{refcount} 
\usepackage{upquote}  

\usepackage{parskip}
\usepackage{amsfonts}
\usepackage{adjustbox}
\usepackage{amsmath}
\usepackage{csquotes}
\usepackage{mdframed}
\usepackage{wrapfig}
\usepackage{array} 
\usepackage{xcolor}
\usepackage[ruled]{algorithm2e}
\usepackage{mathrsfs}
\MakeOuterQuote{"}
\usepackage{makecell}
\usepackage{tikz}
\usetikzlibrary{positioning}
\usepackage{verbatim}
\usepackage{changepage}      
\usepackage{float}
\usepackage{enumitem}
\usepackage{bm}
\usepackage{setspace}
\usepackage{caption}
\usepackage{ifthen}

\usepackage{subcaption}

\DeclareMathOperator*{\argmin}{arg\,min}

\newtheorem{assumption}{Assumption}[section]

\title[Path Integral Optimiser]{Path Integral Optimiser: Global Optimisation via Neural Schr\"odinger-F\"ollmer Diffusion}

\optauthor{%
\Name{Max McGuinness} \Email{mgm52@cam.ac.uk}\\
\Name{Eirik Fladmark} \Email{ef454@cam.ac.uk}\\
\Name{Francisco Vargas} \Email{fav25@cam.ac.uk}\\
\addr University of Cambridge}

\begin{document}

\maketitle

\begin{abstract}%
We present an early investigation into the use of neural diffusion processes for global optimisation, focusing on Zhang et al.'s Path Integral Sampler. One can use the Boltzmann distribution to formulate optimization as solving a Schrödinger bridge sampling problem, then apply Girsanov's theorem with a simple (single-point) prior to frame it in stochastic control terms, and compute the solution's integral terms via a neural approximation (a Fourier MLP). We provide theoretical bounds for this optimiser, results on toy optimisation tasks, and a summary of the stochastic theory motivating the model. Ultimately, we found the optimiser to display promising per-step performance at optimisation tasks between 2 and 1,247 dimensions, but struggle to explore higher-dimensional spaces when faced with a 15.9k parameter model, indicating a need for work on adaptation in such environments. 

\end{abstract}


\section{Introduction}

\textbf{Motivation: \enspace} Most tasks in machine learning can be characterised as \textit{high-dimensional non-convex stochastic optimisation problems}. Stochastic gradient descent (SGD)~\cite{robbins1951SGD} is the archetypical approach in differentiable settings but neglects second-order gradient information. Many extensions have been proposed to incorporate such information---such as Adam~\cite{kingma2014adam}, Adagrad~\cite{duchi2011adagrad}, and momentum~\mbox{\cite{nesterov1983momentum1,tseng1998momentum2}}. However, while effective in the early stages of training, these approaches struggle to generalise as broadly as SGD~\cite{keskar2017adamisbad}, and their theoretical convergence properties remain poorly understood for non-convex global minimisation~\cite{de2018adamconvergence, li2019adagradconvergence}. With diffusion models being shown to be highly effective at sampling from high-dimensional structured distributions~\cite{ramesh2022dalle2}, and offering promising theoretical guarantees~\mbox{\cite{zhang2021pis, vargas2021bayesianneuralSFP}}, this paper wishes to apply those benefits to optimisation.

\textbf{Contributions \enspace} We propose one of the first practical diffusion-based optimisers, the \textit{Path Integral Optimiser (PIO)}: a neural Schr\"odinger-F\"ollmer diffusion process trained as a HyperNetwork. We outline PIO in \textbf{Section~\ref{sec:pio}}; derive theoretical guarantees for PIO as an optimiser in \textbf{Section~\ref{sec:theoretical-guarantees-pio}}; perform an empirical study comparing PIO and PIO-like models to popular optimisers at classic machine learning tasks, in some cases matching or exceeding their performance, in \textbf{Section~\ref{sec:experiments}}; and conclude with recommendations for future work in \textbf{Section~\ref{sec:conclusion}}. We also publish our codebase.\footnote{https://github.com/mgm52/Path-Integral-Optimiser}

\section{Preliminary}

\subsection{The Schr\"odinger-F\"ollmer Proycess} \label{sec:SFP}
The Schr\"odinger bridge problem is that of finding the most likely stochastic evolution from an initial distribution $\mathbb{Q}$ into a target $\mathbb{P}$ while minimising KL-divergence to a Wiener process $W$. We use $P_{W_t}$ to mean the path measure of $W$ at time $t$.
\begin{definition} \label{def:dynamic-SBP-SC-follmer}
    In \textbf{stochastic control terms}, the \textbf{explicit dynamic Schr\"odinger bridge problem} for probability measures $\mathbb{Q}: \mathbb{R}^n \rightarrow \mathbb{R}$ and $\mathbb{P}: \mathbb{R}^n \rightarrow \mathbb{R}$ is that of finding the minimal drift $\mu$ for an identity-variance It\^o process $X_0 \sim \mathbb{Q}$, $X_1 = X_0 + \int_0^1 \mu_t dt + \int_0^1 dW_t$ to transform $\mathbb{Q}$ into $\mathbb{P}$: \vspace{-0.5em}
\begin{align*}
    & \mu = \argmin_{\mu} \mathop{\mathbb{E}}\left[ \frac{1}{2} \int_0^1 \lVert \mu_t \rVert^2 dt + \log \frac{P_{W_1}(X_1)}{\mathbb{P}(X_1)} \right]. 
\end{align*}
\end{definition}

\vspace{-0.5em}
When $\mathbb{Q}$ is a dirac distribution, one can solve this problem via Schr\"odinger-F\"ollmer diffusion~\cite{follmer1984SFP, huang2021schrodinger-follmer-sampler}.

\begin{definition} \label{def:analytic-sfp}
The \textbf{Schr\"odinger-F\"ollmer process} is an identity-variance diffusion process that solves the dynamic Schr\"odinger-F\"ollmer problem from Dirac $X_0 \sim \delta_0 = \mathbb{Q}$ to arbitrary target $X_1 \sim \mathbb{P}$. Taking $f: \mathbb{R}^n \rightarrow \mathbb{R}$ to be the ratio $f(x)=\frac{d \mathbb{P}}{d N\left(0, \mathbf{I}_n\right)}(x)$: \vspace{-0.5em}
\begin{align*}
    X_1 = X_0 + \int_0^1 \frac{\mathbb{E}_{Z \sim N(0, I_n)}[\nabla f(X_t+\sqrt{1-t}  Z)]}{\left.\mathbb{E}_{Z \sim N(0, I_n)}[f(X_t+\sqrt{1-t}  Z)\right.]}dt + \int_0^1 dW_t.
\end{align*}
\end{definition}

\subsection{Motivating a Neural Approximation} \label{sec:mc-vs-neural-sfp}
To overcome the drift's expectation terms not having an analytic closed-form (except in specific cases, like the target being a Gaussian mixture), Huang et al.~\cite{huang2021schrodinger-follmer-sampler} proposed a Monte-Carlo approach. However, Zhang et al.~\cite{zhang2021pis} highlight that this approach is analogous to importance sampling\footnote{To be specific, it may be more apt to compare the SF process to \textit{annealed} importance sampling, which transforms an initial distribution into a target distribution by interpolating the geometric average of the two distributions.}
of target distribution $X_1 \sim \mathbb{P}$ with proposal distribution $Z_i \sim N(0, I_n)$, which entails two major drawbacks encompassing excessively high variance: the \textbf{curse of dimensionality} and the \textbf{curse of distribution alignment} (required samples increases exponentially with KL divergence).

Tzen et al.~\cite{tzen2019theoretical-guarantees-NN-for-SFS} found that, as an alternative to Monte-Carlo estimation, one can use an MLP to efficiently approximate the drift term to arbitrary accuracy, so long as the network can efficiently approximate target density $\mathbb{P}$. Intuitively, a well-formed neural network can increase in depth \textit{linearly} to increase expressivity exponentially~\cite{raghu2017nn-expressive-exp}. We note an additional advantage of the neural/meta approach: no fixed time commitment. One can pause or extend the approximation network's training at any point and then take a viable trajectory; by contrast, the Monte-Carlo method spends all compute running one trajectory to exactly $t=1$, no shorter or longer.


\subsection{Path Integral Sampler}
Zhang et al.~\cite{zhang2021pis} describe a complete implementation of neural-approximated discrete SFP in the \textbf{Path Integral Sampler (PIS)}. PIS learns neural parameters $\theta$ by simply re-using the drift goal in Definition~\ref{def:dynamic-SBP-SC-follmer} as the loss in a standard gradient-based optimisation algorithm (e.g. Adam~\cite{kingma2014adam}).

\begin{definition} \label{def:discrete-neural-SFP}
    The \textbf{EM-discretised neural approximation} to the \textbf{Schr\"odinger-F\"ollmer process} makes use of neural drift term $\widehat{b}_\theta : \mathbb{R}^n \times [0, 1] \rightarrow \mathbb{R}^n$, such that $\widehat{b}$ is a feed-forward multilayer neural net with parameters $\theta \in \mathbb{R}^p$, and is defined for $T$ timesteps, with timesteps $t_i$ = $i/T$, as: \vspace{-0.5em}
\begin{align*}
    X_1 = X_0 + \sum_{i=0}^T \widehat{b}_{\theta}(X_{t_i}, t_i)/T + \sum_{i=0}^T (W_{t_{i+1}}-W_{t_i}).   \vspace{-1em}
\end{align*}
\end{definition}


\begin{definition} \label{def:pis-loss}
    The \textbf{Path Integral Sampler} approximates optimal parameters $\theta^*$ to solve the explicit dynamic Schrodinger Bridge problem \ref{def:dynamic-SBP-SC-follmer} using the \textbf{loss function} goal $\argmin_\theta \mathcal{L}(\theta)\approx\theta^*$, with $\mathcal{L} : \mathbb{R}^p \rightarrow \mathbb{R}$ defined as: \vspace{-0.5em}
\begin{equation*}
    \mathcal{L}(\theta) = \frac{1}{2} \int_0^1 \lVert \widehat{b}_{\theta}(X_t, t) \rVert^2 dt + \log \frac{P_{W_1}(X_1)}{\mathbb{P}(X_1)}. \vspace{-1em}
\end{equation*}
\end{definition}

The assumption that target distribution $\mathbb{P}$ has an analytic form in the loss term characterises the process as solving the \textit{uniform-density-to-$\mathbb{P}$-density sampling problem}, hence Zhang et al. referring to it as a sampler. To enable unbiased sampling, Zhang et al.~\cite{zhang2021pis} perform importance sampling, assigning each trajectory a weight measuring the alignment of its terminal state with $\mathbb{P}$. 

\subsection{SFP For Optimisation} \label{sec:SFP-for-optimization}

One can use a Boltzmann transformation to characterise optimisation as a \textit{sampling} problem.

\begin{definition} \label{def:opt-sampler-transformation}
Given loss function $V : \mathbb{R}^n \rightarrow \mathbb{R}$ and scalar $\sigma \in (0, 1]$, we define the \textbf{Boltzmann density} $\mathbb{P}$ as
    \begin{equation*}
        \mathbb{P}(\phi) = \frac
            {\exp(-V(\phi)/\sigma)}
            {\int_{\mathbb{R}^n} \exp (-V(y)/\sigma)dy}.
    \end{equation*}
For sufficiently small $\sigma$, sampling from $\mathbb{P}$ is equivalent to identifying the optimal parameters $\phi^* \in \argmin_{\phi \in \mathbb{R}^n} V(\phi)$ such that $\lim_{\sigma \rightarrow 0} \mathbb{P}(\phi^*) = 1$.
\end{definition}

Because Schr\"odinger-F\"ollmer processes (\ref{def:analytic-sfp}) are scale-invariant with regard to the target distribution---i.e. constant factors in $f$ are cancelled out by the $\nabla f / f$ division in the drift term---one can ignore the Boltzmann distribution's integral denominator $\int_{\mathbb{R}^n} \exp (-V(y)/\sigma)dy$ and sample with the more readily-computable formula $\mathbb{P}(\phi) = \exp(-V(\phi)/\sigma)$.

To enable the drift term to appropriately match the scale dictated by $\sigma$, Dai et al.~\cite{dai2021global_opt} rescaled the process's terms using $\sigma$ and $\sqrt{\sigma}$: $X_1 = X_0 + \int_0^1 \sigma \widehat{b}_{\theta}(X_t, t) dt + \int_0^1 \sqrt{\sigma} dW_t$.\footnote{This rescaling is not justified in the paper; it appears that Dai et al. may have been motivated by the rescaling that is necessary for \textit{Langevin} dynamics, though we are not certain that the same is necessary for SFP.} Simulating the process targeting a Carrillo function, they found the sampler to be successful at global optimisation, surpassing the Langevin-dynamics sampling method. However, while Dai et al.'s results on the Carrillo function are promising, there has not yet been an attempt to use an SFP for stochastic optimisation, nor machine-learning optimisation, which presents scaling "curses" for the Monte-Carlo approach as mentioned in Section \ref{sec:mc-vs-neural-sfp}.

\subsection{Alternative Processes} \label{alt-processes}
While this paper focuses on PIS as an archetype, it can be seen as part of a greater framework of methods that solve the Schr\"odinger Bridge problem and can thus be applied to optimisation---with varying theoretical guarantees. Notable alternatives include \textbf{Langevin dynamics} of the form $dX_T = -\nabla \frac{1}{2}\nabla \log f(X_t, t)dt + dW_t$ for target density $f$, which tend to require a far longer mixing time than SFP \cite{dai2021global_opt}; and \textbf{denoising diffusion} with a time-reversal $dY_t = -\beta_{T-t}(Y_t + 2\sigma^2 \nabla \log p_{T-t}(Y_t))dt + \sigma \sqrt{2\beta_{T-t}}dW_t$ from $Y_0 \sim \mathcal{N}_{\text{approx}}(0, \sigma^2)$, which may be more numerically stable than SFP, but with additional approximation error due to starting from a gaussian \cite{vargas2023dds}.

Sampling-based optimization methods that \textit{don't} require loss gradient information (PIO uses $d\mathcal{L}/d\theta$) include \textbf{simulated annealing} (adjusts prior with decreasing temperature) \cite{kirkpatrick1983simulatedannealing}, the \textbf{cross-entropy method} (adjusts prior distribution towards N-lowest-loss samples) \cite{RUBINSTEIN1997crossentropymethod}, and \textbf{model predictive path integral control} (MPPI) (each trajectory step executes control on a system, e.g. driving) \cite{theodorou2010mppi}.

\section{Path Integral Optimiser} \label{sec:pio}

\begin{figure}[H]
\centering
\includegraphics[width=0.85\linewidth]{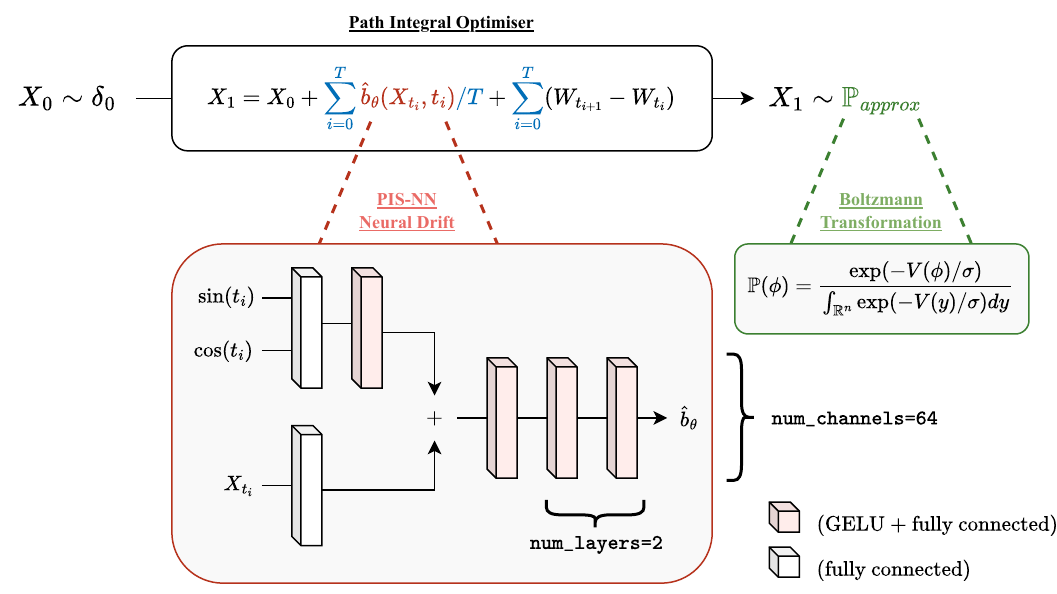} 
\caption{\small The \textbf{Path Integral Optimiser} is a neural-approximated Schr\"odinger-F\"ollmer process (\ref{def:discrete-neural-SFP}) which minimises a loss function $V$ by learning to generate samples from its Boltzmann distribution (green). The drift term is computed by a Fourier MLP (red), and the process as a whole is simulated with Euler-Maruyama discretisation (blue). With correct parameterisation of $\theta, \sigma$, and $T$, this architecture is capable of global optimisation as proven in this paper's corollary (\ref{co:pbigo}).}
\label{fig:pio}
\end{figure}

To perform practical optimisation via a Schr\"odinger-F\"ollmer process, alleviating the scaling issues of Dai et al.'s Monte-Carlo optimiser (Section \ref{sec:SFP-for-optimization}), we adapt Xiao et al.'s Path Integral Sampler~\cite{zhang2021pis} architecture to sample from the Boltzmann transformation of a parameters-to-loss function. We refer to this complete model as a \textbf{Path Integral Optimiser} (PIO). Each training step involves computing at least one full trajectory, then Adam-updating PIO's approximation network according to PIS loss (\ref{def:pis-loss}). At validation, we run multiple trajectories and take the argmin validation loss. An unusual advantage of PIO as an optimizer is that, once trained, we effectively get an ensemble of variants for "free" by computing new stochastic trajectories.

Architecturally, PIO uses the \textit{PIS-NN} approach put forward by Zhang et al.~\cite{zhang2021pis}, in which a single time-conditioned feed-forward network $NN_\theta(X_t, t)$ approximates the entire drift. Alternatively, PIO-Grad uses the \textit{PIS-Grad} formulation, which may be less impractical for general machine learning optimisation due to the cost of re-computing target score $\nabla \log \mathbb{P}$ at each timestep as guidance. To assist PIO at optimisation, we decrease the Boltzmann parameter $\sigma$ in lockstep with learning rate, matching higher-variance loss landscapes with finer-grained parameter updates.



\section{Theoretical Guarantees} \label{sec:theoretical-guarantees-pio}
We derive a core theorem for the usability of discretised \textit{neural-approximated} Schr\"odinger-F\"ollmer-like processes (\ref{def:discrete-neural-SFP}) such as PIO for global optimisation, by building on the theoretical results of several SFP papers~\cite{tzen2019theoretical-guarantees-NN-for-SFS,dai2021global_opt,vargas2021bayesianneuralSFP}. These bounds are also relevant to \textit{denoising diffusion} optimizers (\ref{alt-processes}), which one may call F\"ollmer-like due to satisfying the same expressiveness bounds and having a drift term that can be similarly expressed in terms of a semigroup (the semigroup being defined by score)~\cite{smooth-cloud-dds-theory-like-sfp}.

To improve interpretability, we provide more informal corollaries in the \textit{Remarks} section. Full derivations can be found in the appendix. 

\begin{theorem} \label{th:pwithconstants}
    Suppose assumptions \ref{as:A} - \ref{as:t3.1} hold.  Then, for each $\epsilon \in (0, \tau), \sigma \in (0, 1), \widehat{\epsilon} \in (0, 16L^2/c^2), T \in \mathbb{Z}^+$, there exist constants $C_{\tau,\epsilon,n}$ and $C_{L'}$ (depending on $\tau, \epsilon, n$ and $L'$ respectively) such that \vspace{-1em}
    $$
    P_{\widehat{Y}_1}(V(X_1) > \tau) \leq C_{\tau,\epsilon,n}\exp\left(-\frac{\tau-\epsilon}{\sigma}\right) + \sqrt{2\widehat{\epsilon}} + \sqrt{4{L'}^2 (C_{L'}/T + 1/T^2)},\vspace{-1em}
    $$ 
    for Schr\"odinger F\"ollmer process $\widehat{Y}$ EM-discretised with step size $1/T$, whose neural drift $\widehat{b}_\theta(x, t)=\widehat{v}(x, \sqrt{1-t})$ is $L'$-Lipschitz in both parameters.
\end{theorem}

\subsubsection{Remarks}
Theorem \ref{th:pwithconstants} may be more intuitively phrased as \textit{the probability of discrete neural SFP $\widehat{Y}$ being a $\tau$-global minimiser of function $V$:}\vspace{-0.5em}
\begin{align*}
P_{\widehat{Y}_1}(V(X_1) < \tau) \geq 1 - \left(C_{\tau,\epsilon,n}\exp\left(-\frac{\tau-\epsilon}{\sigma}\right) + \sqrt{2\widehat{\epsilon}} + \sqrt{4{L'}^2 (C_{L'}/T + 1/T^2)}\right).\vspace{-1em}
\end{align*}
We can trivially see that Theorem \ref{th:pwithconstants} enables global optimisation by considering the fact that, by Dai et al.'s theorem, there exists a neural SFP drift for every $\widehat{\epsilon}>0$; that by Tzen et al.'s assumptions, we can set $\sigma \in (0, 1]$ arbitrarily; and that by EM-discretisation, we can set $1/T \in (0, 1]$ arbitrarily.

Therefore, the three parameters can tend to 0 to reach a global optimiser:
\begin{align} \notag
& \lim_{\sigma, \widehat{\epsilon}, 1/T \rightarrow 0} \left[ C_{\tau,\epsilon,n}\exp\left(-\frac{\tau-\epsilon}{\sigma}\right) + \sqrt{2\widehat{\epsilon}} + \sqrt{4{L'}^2 (C_{L'}/T + 1/T^2)} \right] \\ \notag
= \; & \lim_{\sigma \rightarrow 0} C_{\tau,\epsilon,n}\exp\left(-\frac{\tau-\epsilon}{\sigma}\right) + \lim_{\widehat{\epsilon} \rightarrow 0} \sqrt{2\widehat{\epsilon}} + \lim_{1/T \rightarrow 0} \sqrt{4{L'}^2 (C_{L'}/T + 1/T^2)} \\ \notag
\rightarrow \; & 0 + 0 + 0. 
\end{align}
Additionally, as a more informal result from Theorem \ref{th:pwithconstants}, we present the following corollary:
\begin{corollary} \label{co:pbigo}
    By Theorem \ref{th:pwithconstants}, $\forall 0<\delta \ll 1$, with probability at least $1~-~\sqrt{\delta}$, $\widehat{Y}_1$ is a $\tau$-global minimiser of $V$, i.e., $V(\widehat{Y}_1) \leq \tau+\inf V(x)$, if for the scaling factor $\sigma$, neural error $\hat{\epsilon}$, and number of iterations $T$, we have:
    $$\mathcal{O} \left(\frac{\tau-\epsilon}{\ln \left(1 / \delta \right)} \right) \geq \sigma
\quad\land\quad \mathcal{O} \left( \delta \right) \geq \widehat{\epsilon}
\quad\land\quad  \mathcal{O} ({L'} / \sqrt{\delta})\leq T.$$
\end{corollary}

\section{Experimental Results} \label{sec:experiments}




\begin{table}[H] 
\centering
\begin{tabular}{|c|c||c|c|c|c|}
\hline
\textbf{Sweep} & \textbf{Seed} & \textbf{Optimiser} & \textbf{Carrillo~\cite{carrillo2021consensus}} & \textbf{Moons~\cite{balfer2017scikit-learn-moons}} & \textbf{MNIST~\cite{lecun1998mnist}} \\
Runs/Steps & Runs/Steps & & ($|\theta|=2$) & ($|\theta|=41$) & ($|\theta|\approx15.9k$) \\
\hline\hline
\multirow{4}{*}{\textbf{32/32}} & \multirow{4}{*}{\textbf{10*/100}} 
& \textbf{Adagrad~\cite{duchi2011adagrad}} & \textit{0.549} $\pm$ 0.153 & 0.184 $\pm$ 0.036 & \textbf{0.425} $\pm$ 0.031 \\
& & \textbf{Adam~\cite{kingma2014adam}} & 0.597 $\pm$ 0.536 & \textbf{0.057} $\pm$ 0.047 & 0.476 $\pm$ 0.030 \\
& & \textbf{SGD~\cite{robbins1951SGD}} & \textbf{0.491} $\pm$ 0.125 & 0.356 $\pm$ 0.019 & \textit{0.427} $\pm$ 0.034 \\
& & \textbf{PIO} & 1.458 $\pm$ 0.882 & \textit{0.129} $\pm$ 0.121 & 2.464 $\pm$ 0.026 \\
\hline
\multirow{4}{*}{\textbf{64/64}} & \multirow{4}{*}{\textbf{10*/100}} 
& \textbf{Adagrad~\cite{duchi2011adagrad}} & 0.647 $\pm$ 0.448 & 0.193 $\pm$ 0.038 & \textit{0.347} $\pm$ 0.020 \\
& & \textbf{Adam~\cite{kingma2014adam}} & \textit{0.498} $\pm$ 0.000 & \textbf{0.022} $\pm$ 0.010 & \textbf{0.346} $\pm$ 0.020 \\
& & \textbf{SGD~\cite{robbins1951SGD}} & 0.553 $\pm$ 0.149 & 0.428 $\pm$ 0.026 & 0.385 $\pm$ 0.011 \\
& & \textbf{PIO} & \textbf{0.404} $\pm$ 0.417 & \textit{0.032} $\pm$ 0.049 & 2.424 $\pm$ 0.045 \\
\hline
\end{tabular}
\caption{\footnotesize{Min test loss $V(x)$ across optimisers on \textit{mini-batch} training. Bold indicates best-in-column; italics second-best. The 64/64 runs make use of an annealing routine that drops both LR and PIO's $\sigma$ parameter by $50\%$ on plateaus. *5 runs for MNIST, 10 otherwise.} }
\label{tab:main-results}
\end{table}
\vspace{-1.5em}
As a work-in-progress result, in Table \ref{table:finaleval-updated}, we find PIO-Grad to provide more competitive per-step performance at the cost of compute (of around 10$\times$ that of PIO for Circles vs Moons). We also compare to DDO, a denoising diffusion sampler~\cite{vargas2023dds} (\ref{alt-processes}) driven by a PIS-Grad-like network.\footnote{Results from an independent implementation to Table \ref{tab:main-results}, using a different codebase and experimental setup.}

\begin{table}[H] 
\centering
\renewcommand{\arraystretch}{0.85} 
\begin{tabular}{|c|c||c|c|c|c|} 
\hline
\textbf{Sweep} & \textbf{Seed} & \textbf{Optimiser} & \textbf{Circles~\cite{balfer2017scikit-learn-moons}} & \textbf{Breast Cancer~\cite{Dua:2019}} & \textbf{Covertype~\cite{Dua:2019}} \\
Runs/Steps & Runs/Steps & & ($|\theta|=64$) & ($|\theta|=171$) & ($|\theta|=1,247$) \\
\hline\hline
\multirow{5}{*}{\textbf{6/300*}} & \multirow{5}{*}{\textbf{1/300*}} 
& \textbf{Adam~\cite{kingma2014adam}} & \textit{0.998} & 0.954 & 0.772 \\
& & \textbf{AdamW~\cite{adamw-2019}} & \textit{0.998} & \textit{0.966} & 0.766 \\
& & \textbf{Noisy-SGD~\cite{noisy-sgd}} & 0.997 & 0.941 & 0.684 \\
& & \textbf{SGD~\cite{robbins1951SGD}} & 0.997 & \textbf{0.970} & 0.769 \\
& & \textbf{SGLD~\cite{sgld_opt_2011}} & 0.934 & 0.965 & 0.656 \\
\hline
\multirow{2}{*}{\textbf{3/300*}} & \multirow{2}{*}{\textbf{1/300*}} 
& \textbf{DDO-Grad} & \textbf{0.999} & 0.951 & \textit{0.801} \\
& & \textbf{PIO-Grad} & \textit{0.998} & 0.944 & \textbf{0.813} \\
\hline
\end{tabular}
\caption{\footnotesize{Test accuracy \% across optimisers on \textit{full-batch} training. Bold indicates best-in-column; italics second-best. *150 steps for Covertype, 300 otherwise.}}\label{table:finaleval-updated}
\end{table}

\section{Conclusions} \label{sec:conclusion}



Despite the benefits of being able to stochastically generate alternate parameterisations, the application of diffusion models to ML-optimization has had little attention in literature. We address this by introducing the Path Integral Optimiser (PIO), an application of the Path Integral Sampler to optimisation by use of the Boltzmann transformation.  We derive theoretical guarantees for such discretized neural Schr\"odinger-F\"ollmer processes at the task of optimization, showing that---for example---the parameter $\sigma$ should decrease logarithmically with $\delta$ to achieve P(global optimisation) over $1 - \sqrt{\delta}$. In empirically evaluating PIO, we match or exceeds classical optimizers in tasks up to 1.5k parameters in size, but the implementation struggles to scale when given an 15.9k-parameter MLP MNIST optimisation problem. Future work should focus on improving scalability by considering larger drift-approximation networks, ensembling PIO's trajectories, pre-training to a known distribution, and better-parallelising the training routine to make PIS-Grad more performant.


\bibliography{main}
\newpage
\clearpage
\appendix

\section{Proof of Main Theorem \ref{th:pwithconstants}} \label{sec:proof-section}
\subsection{General Approach}
We can build up our result by considering several Schr\"odinger-F\"ollmer processes:
\begin{itemize}
    \item $X$: A continuous analytic SFP (\ref{def:analytic-sfp}).
    \item $\widehat{X}$: A continuous neural SFP (\ref{def:discrete-neural-SFP} but with integrals instead of EM-discretisation).
    \item $\widehat{Y}$: A discrete neural SFP (\ref{def:discrete-neural-SFP}).
\end{itemize}

To evaluate $\widehat{Y}$'s proficiency at global optimisation for a given loss function $V$, the aim is to find a tight upper bound on the probability of $V(\widehat{Y}_1)>\tau$. In other words, to find the probability of our optimiser finding a solution with loss worse than $\tau$.

Using the pushforward measure $P_{\widehat{Y}_1}$, one can write this as\footnote{Note that this takes advantage of the fact that $V(X_1)$ and $V(\widehat{Y}_1)$ have the same event space, thus $V(X_1)>\tau$ is the same event as $V(\widehat{Y}_1)>\tau$.} $$P_{\widehat{Y}_1}(V(X_1) > \tau).$$

To consider $\widehat{Y}$ a global optimiser, it should be possible to reduce this probabiliy to near-0 for any $\tau$ with correct choice of neural drift parameters $\theta$ as used in Definition \ref{def:discrete-neural-SFP}.

\subsection{Assumptions}
To make use of the results of prior authors, we will first conglomerate their assumptions. To begin, to ensure that the SDE for $X$ admits a unique and strong solution, Dai et al.~\cite{dai2021global_opt} assume:
\begin{assumption} \label{as:A}
    $V(x)$ is twice continuous differentiable on $\mathbb{R}^n$ and $V(x) = \lVert x \rVert_2^2/2$ outside a ball $B_R$, where $B_R$ denotes the ball centred at origin with radius $R > 0$.
\end{assumption}
Additionally, Tzen et al. \cite{tzen2019theoretical-guarantees-NN-for-SFS} make several assumptions to ensure that the ratio $f$ used in the analytic SFP definition \ref{def:analytic-sfp} can be efficiently approximated by a neural net:
\begin{assumption} \label{as:3.1}
    For the analytic SFP definition (\ref{def:analytic-sfp}), $f$ is differentiable, $f$ and $\nabla f$ are $L$-Lipschitz, and there exists a minima $c \in (0, 1]$ such that $f \geq c$ everywhere.
\end{assumption}
\begin{assumption} \label{as:3.3}
    For any $R>0$ and $\hat{\epsilon}>0$, there exists a feedforward neural net $\widehat{f}$ with size polynomial in $(1 / \hat{\epsilon}, n, L, R)$, such that
    $$
    \sup _{x \in \mathrm{B}^n(R)}|f(x)-\widehat{f}(x)| \leq \hat{\epsilon} \quad \text { and } \sup _{x \in \mathrm{B}^n(R)}\|\nabla f(x)-\nabla \widehat{f}(x)\| \leq \hat{\epsilon}.
    $$
\end{assumption}
\begin{assumption} \label{as:t3.1}
     For neural SFP $\widehat{X}$, the drift $\widehat{b}_\theta(x, t)=\widehat{v}(x, \sqrt{1-t})$ is determined by a neural net $\widehat{v}:\mathbb{R}^n \times[0,1] \rightarrow \mathbb{R}^n$ with size polynomial in $(1 / \hat{\epsilon}, n, L, c, 1 / c)$, such that the activation function of each neuron is an element of the set $\left\{\operatorname{S}, \operatorname{S}^{\prime}, \operatorname{ReLU}\right\}$, where $\operatorname{S}$ is the sigmoid function.
\end{assumption}
It is worth noting that none of our assumptions require $V$ to be convex within the ball $B_R$; thus the global optimisation problem remains challenging.

\subsection{Proof of Main Theorem \ref{th:pwithconstants}}

We can phrase our optimisation probability in terms of its distance from the continuous neural SFP terminal distribution $\widehat{X}_1$:
\begin{align} \notag
P_{\widehat{Y}_1}(V(X_1) > \tau) &= P_{\widehat{X}_1}(V(X_1) > \tau) + P_{\widehat{Y}_1}(V(X_1) > \tau) - P_{\widehat{X}_1}(V(X_1) > \tau) \\
&\leq P_{\widehat{X}_1}(V(X_1) > \tau) + |P_{\widehat{Y}_1}(V(X_1) > \tau) - P_{\widehat{X}_1}(V(X_1) > \tau)|. \label{eq:p1}
\end{align}
Next, consider the total variation norm as given in Definition \ref{def:tv-distance}.
\begin{definition} \label{def:tv-distance}
    For probability measures $P$ and $Q$ in measurable space $(\Omega, \mathcal{F})$, the total variation norm is defined as $2\times$ the upper bound on the distance between $P$ and $Q$:
$$\lVert P - Q \rVert_{TV} = 2 \sup_{A \in \mathcal{F}} |P(A) - Q(A)|.$$
\end{definition}
Naturally, as the total variation norm is an upper bound, we have $|P(A) - Q(A)| = |Q(A) - P(A)| \leq \lVert Q - P \rVert_{TV}$ for all events $A \in \mathcal{F}$. Considering $(V(X_1) > \tau)$ as an event in the event space of the probability measures for $\widehat{X}_1$ and $\widehat{Y}_1$, we can apply the bound to Equation \ref{eq:p1}, giving us:
\begin{align} \notag
&P_{\widehat{X}_1}(V(X_1) > \tau) + |P_{\widehat{Y}_1}(V(X_1) > \tau) - P_{\widehat{X}_1}(V(X_1) > \tau)| \\ 
\leq \; & P_{\widehat{X}_1}(V(X_1) > \tau) + \lVert P_{\widehat{X}_1} - P_{\widehat{Y}_1} \rVert_{TV}.
\end{align}
We can then move from the neural approximation $\widehat{X}$ to the analytic process $X$ using the same trick:
\begin{align} \notag
& P_{\widehat{X}_1}(V(X_1) > \tau) + \lVert P_{\widehat{X}_1} - P_{\widehat{Y}_1} \rVert_{TV} \\
\leq \; & P_{X_1}(V(X_1) > \tau) + \lVert P_{X_1} - P_{\widehat{X}_1} \rVert_{TV} + \lVert P_{\widehat{X}_1} - P_{\widehat{Y}_1} \rVert_{TV}.
\end{align}
Next, we can obtain a KL divergence representation by applying Pinsker's inequality, which I define in \ref{def:pinsker}.
\begin{definition} \label{def:pinsker}
    For probability measures P and Q, Pinsker's inequality states that
    $$\lVert P - Q \rVert_{TV} \leq \sqrt{2D_{KL}(P \parallel Q)}.$$
\end{definition}
Thus, by Pinsker's inequality, we have:
\begin{align} \notag
& P_{X_1}(V(X_1) > \tau) + \lVert P_{X_1} - P_{\widehat{X}_1} \rVert_{TV} + \lVert P_{\widehat{X}_1} - P_{\widehat{Y}_1} \rVert_{TV} \\
\leq \; & P_{X_1}(V(X_1) > \tau) + \sqrt{2D_{KL}(P_{X_1} \parallel P_{\widehat{X}_1})} + \sqrt{2D_{KL}(P_{\widehat{X}_1} \parallel P_{\widehat{Y}_1})}. \label{eq:dkl-bounds}
\end{align}
To continue, we will take assumptions \ref{as:A} - \ref{as:t3.1} and apply three results from prior authors:
\begin{itemize}
    \item \textit{\underline{Dai et al.}} \cite{dai2021global_opt} (Analytic optimisation bound 3.5): \\$P_{X_1}(V(X_1)>\tau)\leq~C_{\tau,\epsilon,n}\exp\left(-\frac{\tau-\epsilon}{\sigma}\right)$ under assumption \ref{as:A}.
    \item \textit{\underline{Tzen et al.}} \cite{tzen2019theoretical-guarantees-NN-for-SFS} (Neural approximation bound 3.1):
    \\$D_{KL}(P_{X_1} \parallel P_{\widehat{X}_1}) \leq \widehat{\epsilon}$ under assumptions \ref{as:3.1} - \ref{as:t3.1}.
    \item \textit{\underline{Vargas et al.}} \cite{vargas2021bayesianneuralSFP} (EM-Discretisation bound A6):
    \\$D_{KL}(P_{\widehat{X}_1} \parallel P_{\widehat{Y}_1}) \leq 2{L'}^2 (C_{L'}/T + 1/T^2)$ under assumptions \ref{as:3.1} - \ref{as:t3.1}.
\end{itemize}
Inserting these into Equation \ref{eq:dkl-bounds}, we obtain:
\begin{align} \notag
& P_{X_1}(V(X_1) > \tau) + \sqrt{2D_{KL}(P_{X_1} \parallel P_{\widehat{X}_1})} + \sqrt{2D_{KL}(P_{\widehat{X}_1} \parallel P_{\widehat{Y}_1})} \\
\leq \; & C_{\tau,\epsilon,n}\exp\left(-\frac{\tau-\epsilon}{\sigma}\right) + \sqrt{2\widehat{\epsilon}} + \sqrt{4{L'}^2 (C_{L'}/T + 1/T^2)}.
\end{align}
In summary, we have obtained a bound for $\widehat{Y}$ as an optimiser, proving Theorem \ref{th:pwithconstants}:
\begin{align} 
P_{\widehat{Y}_1}(V(X_1) > \tau) \leq C_{\tau,\epsilon,n}\exp\left(-\frac{\tau-\epsilon}{\sigma}\right) + \sqrt{2\widehat{\epsilon}} + \sqrt{4{L'}^2 (C_{L'}/T + 1/T^2)}.
\end{align}

\section{Proof of Main Corollary \ref{co:pbigo}}
We can examine the terms of our core result \ref{th:pwithconstants} in isolation to derive bounds on parameter $\sigma$, neural error $\widehat{\epsilon}$ and step count $K$ for optimisation accuracy. In particular, suppose we wish to achieve $\tau$-global minimisation with probability at least $1 - \sqrt{\delta}$ for arbitrary $\delta$, meaning
\begin{equation} \label{eq:b1}
P_{\widehat{X}_1}(V(X_1) > \tau) \leq C_{\tau,\epsilon,n}\exp\left(-\frac{\tau-\epsilon}{\sigma}\right) + \sqrt{2\widehat{\epsilon}} + \sqrt{4{L'}^2 (C_{L'}/T + 1/T^2)} \leq \sqrt{\delta}.
\end{equation}
We can achieve this by dividing $\sqrt{\delta}$ into three parts using constants $C_1+C_2+C_3 = 1$, such that our bound in Equation \ref{eq:b1} is equivalent to solving the system:
\begin{align} \notag
    C_{\tau,\epsilon,n}\exp\left(-\frac{\tau-\epsilon}{\sigma}\right) \leq~C_1 \sqrt{\delta} \quad \land \quad \sqrt{2\widehat{\epsilon}}\leq~C_2\sqrt{\delta} \quad \land \quad \sqrt{4{L'}^2 (C_{L'}/T + 1/T^2)}\leq~&C_3\sqrt{\delta}.
\end{align}
I will now address each part individually; careful inequality handling can lead each term to a big-$\mathcal{O}$ bound, proving Corollary \ref{co:pbigo}:

\vspace{1cm}

\begin{adjustbox}{center}
\begin{tabular}{p{0.35\textwidth}|p{0.35\textwidth}|p{0.35\textwidth}}
\textbf{1.} Firstly, manipulating the $\sigma$ bound (corresponding to the gap between sampling and optimisation via Definition \ref{def:opt-sampler-transformation}): & \textbf{2.} Secondly, manipulating the $\widehat{\epsilon}$ bound (corresponding to the gap between the analytical SFP and its neural approximation): & \textbf{3.} Finally, manipulating the $T$ bound (corresponding to the gap between continuous and EM-discretised neural SFP): \\

\centering $\!\begin{aligned} \notag
C_{\tau,\epsilon,n}\exp\left(-\frac{\tau-\epsilon}{\sigma}\right) & \leq C_1 \sqrt{\delta} \\ \notag
\ln \left( \frac{C_1}{C_{\tau,\epsilon,n}} \sqrt{\delta} \right) & \geq -\frac{\tau-\epsilon}{\sigma}  \\ \notag
\frac{1}{2} \ln \left( \frac{C_1^2}{C_{\tau,\epsilon,n}^2} \delta \right) \sigma & \geq -(\tau-\epsilon)  \\ \notag
-\frac{2(\tau-\epsilon)}{\ln \left( \frac{C_1^2}{C_{\tau,\epsilon,n}^2} \delta \right)} & \geq \sigma \\ \notag 
\mathcal{O} \left(\frac{\tau-\epsilon}{\ln \left(1 / \delta \right)} \right) & \geq \sigma.
\end{aligned}$

& 

\vspace*{-78pt}
\centering $\!\begin{aligned} \notag
\sqrt{2\widehat{\epsilon}} & \leq C_2 \sqrt{\delta} \\ \notag
\widehat{\epsilon} & \leq C_2^2 \delta / 2 \\
\mathcal{O} \left( \delta \right) & \geq \widehat{\epsilon}.
\end{aligned}$

& 

\vspace*{-80pt}
\centering $\!\begin{aligned} \notag
\sqrt{4{L'}^2 (C_{L'}/T + 1/T^2)}&\leq~C_3\sqrt{\delta} \\ \notag
(4{L'}^2) (C_{L'}/T + 1/T^2)&\leq~C_3^2\delta \\ \notag
(4{L'}^2) (TC_{L'} + 1)&\leq~T^2C_3^2\delta \\ \notag
(4{L'}^2) (C_{L'} + 1)&\leq~T^2C_3^2\delta \\ \notag
\frac{(C_{L'} + 1)(4{L'}^2)}{C_3^2\delta}&\leq~T^2 \\
\mathcal{O}({L'} / \sqrt{\delta})&\leq~T.
\end{aligned}$
\end{tabular}
\end{adjustbox}

\section{Evaluation Details}

In evaluating PIO, we wish to account for the fact that a practical application of the optimiser would involve some degree of hyperparameter tuning; we also account for inherent stochasticity in both the diffusion model and dataset by re-running the optimisation over multiple seeds; and evaluate the optimiser at multiple tasks to assess generalisation.

\subsection{Hyperparameter Sweeps}
Optimisers are considerably challenging to compare fairly---in 2020, Choi et al. found that minor alterations to hyperparameter search space can \textbf{reverse classic optimisation results}, emphasising that hyperparameter relationships should not be neglected \cite{choi2019optimisershyperparamsmisleading}. For example, they found a tendency for researchers to underestimate Adam's performance by not including its full breadth of parameters in a sweep. 


To address this, we consider two distinct sweeps, each performed using the $\texttt{BayesOpt}$ Bayesian optimisation algorithm included in Meta's Nevergrad library \cite{nevergrad}. Firstly, \textbf{Narrow $\times$32}, a 32-run sweep with each run lasting 32 training steps. Secondly, \textbf{Wide $\times$64}: a 64-run sweep, each run lasting 64 training steps and making use of an LR-scheduling algorithm that drops LR by $50\%$ on plateaus, and scales PIO's $\sigma$ parameter to match.

\vspace{1em}

\begin{itemize}
    \item \begin{mdframed}[linewidth=0.6pt, innerleftmargin=18pt, innertopmargin=10pt, innerbottommargin=10pt, innerrightmargin=18pt]
        \textbf{Narrow $\times$32}: a 32-run sweep, each run lasting 32 training steps, over the following hyperparameters:
        \begin{itemize}[noitemsep]
            \item PIO: \texttt{lr, grad\_clip\_val, batch\_size, sigma}.
            \item Others: \texttt{lr, grad\_clip\_val, batch\_size}.
        \end{itemize}
    \end{mdframed}
    \vspace{1em}
    \item \begin{mdframed}[linewidth=0.6pt, innerleftmargin=18pt, innertopmargin=10pt, innerbottommargin=10pt, innerrightmargin=18pt]
        \textbf{Wide $\times$64}: a 64-run sweep, each run lasting 64 training steps and making use of an LR-scheduling algorithm that drops LR by $50\%$ on plateaus, and scales PIO's $\sigma$ parameter to match. This is performed over the following hyperparameters:
        \begin{itemize}[noitemsep]
            \item PIO: \texttt{lr, grad\_clip\_val, batch\_size, batch\_laps, sigma, \\num\_layers, K}.
            \item Adam: \texttt{lr, grad\_clip\_val, batch\_size, batch\_laps, beta\_1, beta\_2}.
            \item Others: \texttt{lr, grad\_clip\_val, batch\_size, batch\_laps}.
        \end{itemize}
    \end{mdframed}
\end{itemize}

The idea with \textbf{Wide $\times$64} is that we aim to fully exploit each optimiser by including a wide variety of parameters. The unusual \texttt{batch\_laps} parameter entails re-using the same mini-batch over multiple consecutive training steps, which may improve numerical stability. Figure \ref{fig:res-params-ex} exemplifies parameters for the \textbf{Narrow $\times$32} sweep.

\section{Relevant concepts}

\subsection{Boltzmann transformation and $\sigma$ reduction} \label{univarite_section}

\begin{figure}[!htb]  
    \centering

    \begin{minipage}{0.3\textwidth}
        \centering
        \includegraphics[width=0.9\linewidth]{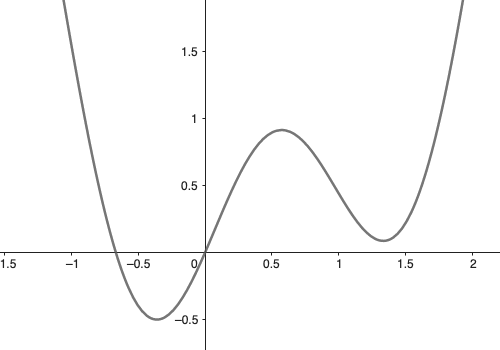}
        \\ \small $V_1(x)$
    \end{minipage}
    \hfill
    \begin{minipage}{0.3\textwidth}
        \centering
        \includegraphics[width=0.9\linewidth]{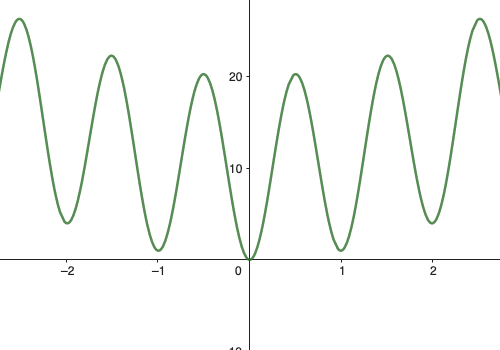}
        \\ \small $V_2(x)$
    \end{minipage}
    \hfill
    \begin{minipage}{0.3\textwidth}
        \centering
        \includegraphics[width=0.9\linewidth]{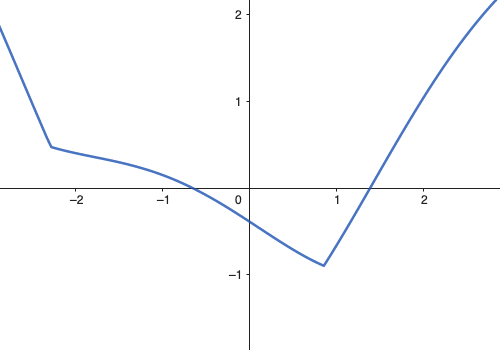}
        \\ \small $V_3(x)$
    \end{minipage}

    \caption{\small Univariate optimisation target functions.}
    \label{univariate}
\end{figure}

\begin{figure}[h!]
    \centering
    
    \begin{minipage}{0.3\textwidth}
        \centering
        \includegraphics[width=0.9\linewidth]{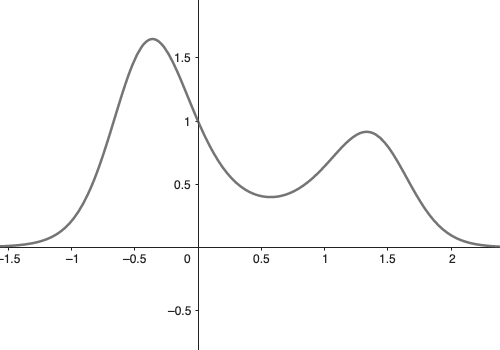}
        \\ \small $e^{\frac{-V_1(x)}{1}}$
    \end{minipage}
    \hfill
    \begin{minipage}{0.3\textwidth}
        \centering
        \includegraphics[width=0.9\linewidth]{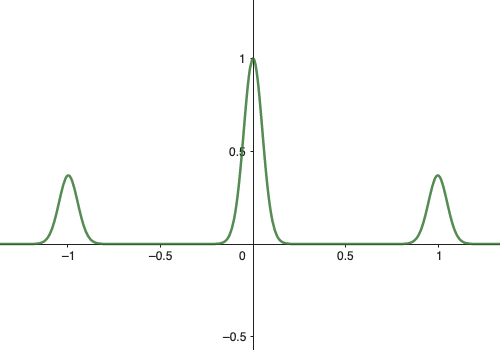}
        \\ \small $e^{\frac{-V_2(x)}{1}}$
    \end{minipage}
    \hfill
    \begin{minipage}{0.3\textwidth}
        \centering
        \includegraphics[width=0.9\linewidth]{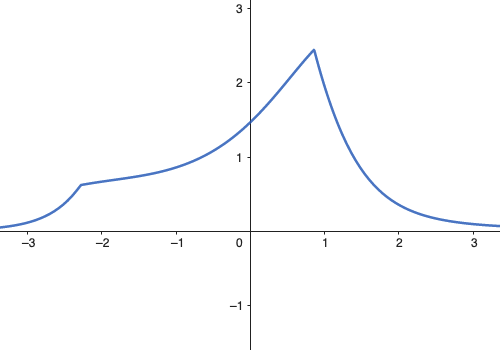}
        \\ \small $e^{\frac{-V_3(x)}{1}}$
    \end{minipage}

    \caption{\small Univariate optimisation target functions following the Boltzman transformation with $\sigma = 1$.}
    \label{univariate_trans}
\end{figure}

\begin{figure}[h!]
    \centering
    
    \begin{minipage}{0.3\textwidth}
        \centering
        \includegraphics[width=0.9\linewidth]{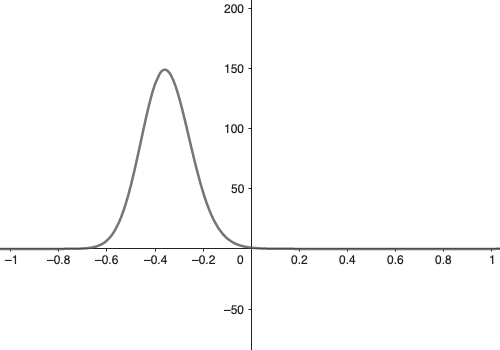}
        \\ \small $e^{\frac{-V_1(x)}{0.1}}$
    \end{minipage}
    \hfill
    \begin{minipage}{0.3\textwidth}
        \centering
        \includegraphics[width=0.9\linewidth]{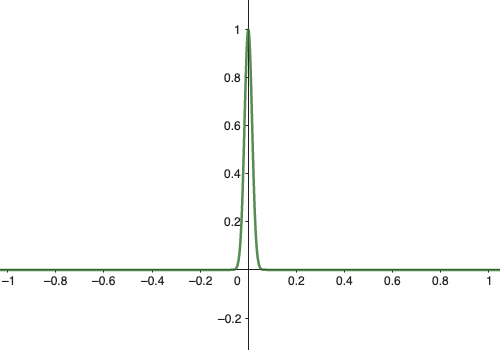}
        \\ \small $e^{\frac{-V_2(x)}{0.1}}$
    \end{minipage}
    \hfill
    \begin{minipage}{0.3\textwidth}
        \centering
        \includegraphics[width=0.9\linewidth]{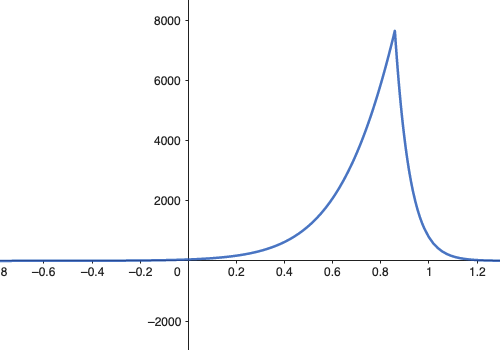}
        \\ \small $e^{\frac{-V_3(x)}{0.1}}$
    \end{minipage}

    \caption{\small Univariate optimisation target functions following the Boltzman transformation with $\sigma = 0.1$.}
    \label{univariate_trans_enhance}
\end{figure}

The following graphics are provided to give a visual representation of how the Boltzman transformation changes a univariate optimisation target (\ref{univariate}), into a Boltzman distribution (\ref{univariate_trans}), and how reducing our optimisation parameter $\sigma$ makes these distributions significantly sharper (\ref{univariate_trans_enhance}).



\clearpage
\section{Further Results}

\subsubsection{Narrow $\times$32 sweep}


\begin{figure}[!htb]
\begin{adjustwidth}{-2cm}{-2cm}  
    \centering
    
    \begin{minipage}{0.45\textwidth}
        \centering
        \includegraphics[width=0.9\linewidth]{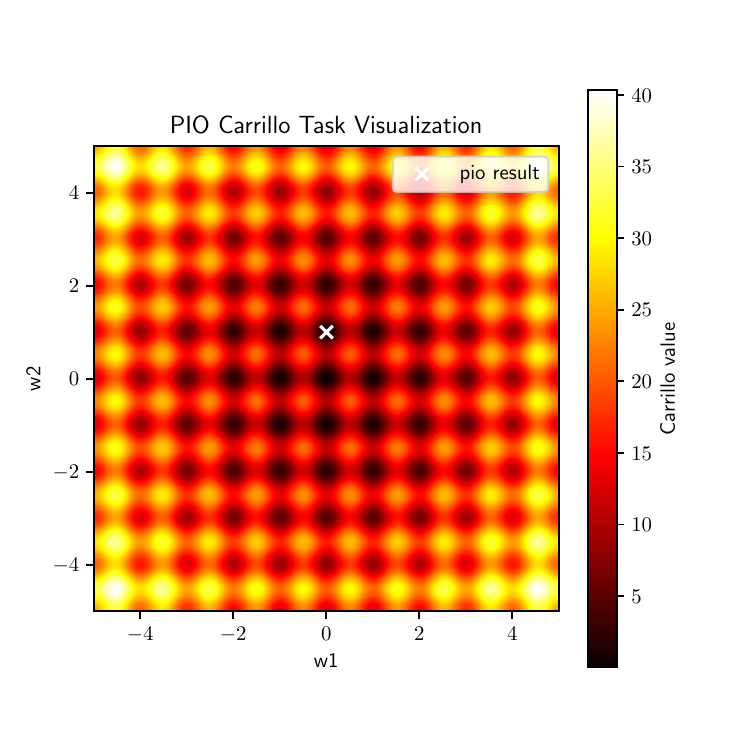}
        \captionsetup{width=0.9\linewidth}
        \caption{\small The Carrillo point found by the best PIO run in its $\times$32 hyperparameter sweep; the optimiser appears stuck in a local minima.}
        \label{fig:res-carrillo-viz}
    \end{minipage}
    \begin{minipage}{0.7\textwidth}
        \centering
        \includegraphics[width=0.45\linewidth]{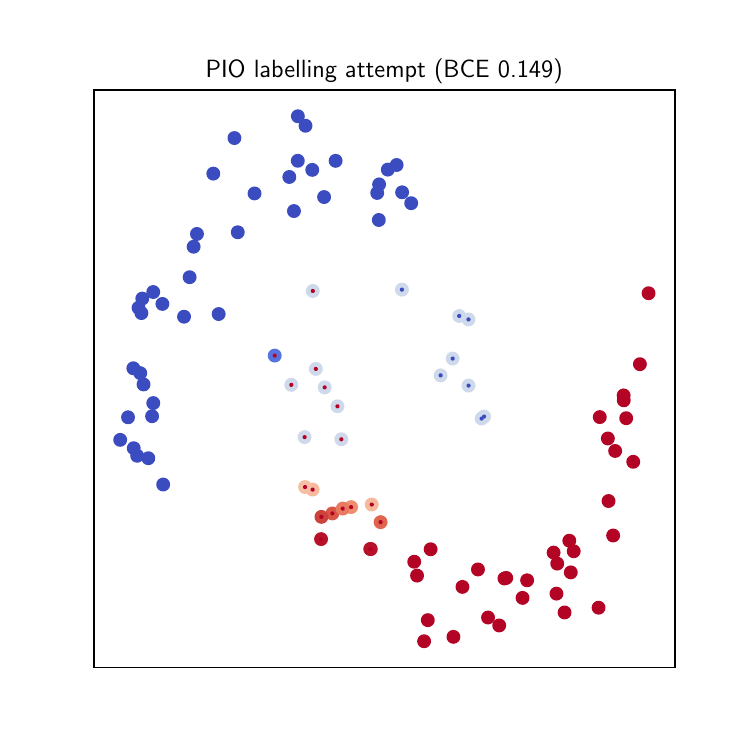}
        \includegraphics[width=0.45\linewidth]{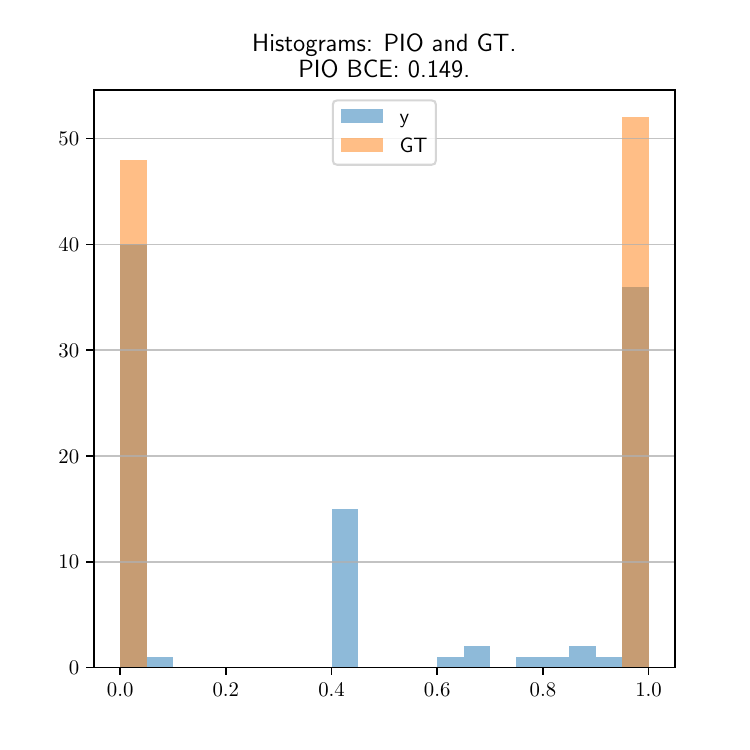} 
        \captionsetup{width=0.9\linewidth}
        \caption{\small On the left: PIO's final attempt at the Moons labelling task after $\times$32 sweeping, with ground truth marked with a small dot in the center of each circle. On the right: a histogram of PIO's labels compared to the ground truth, demonstrating room for improvement.}
        \label{fig:res-moons-viz}
    \end{minipage}
    
\end{adjustwidth}
\end{figure}
\begin{figure}[h]
\begin{adjustwidth}{-3cm}{-3cm}  
    \centering
    
    \begin{minipage}{0.66\textwidth}
        \centering
       \includegraphics[width=0.4\linewidth]{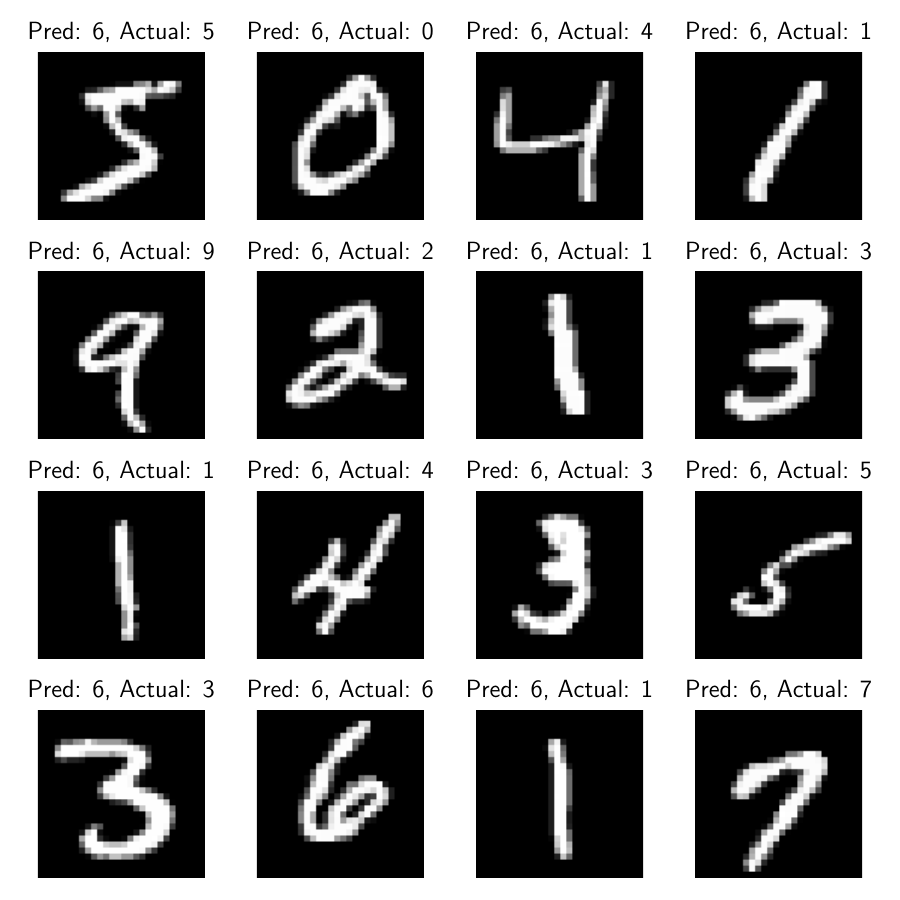} 
       \captionsetup{width=0.8\linewidth}
       \caption{\small The MNIST labels predicted by PIO's best run in its $\times$32 hyperparameter sweep; every label is the same, implying that the optimiser is stuck in a very early local minima.}
       \label{fig:res-mnist-viz}
    \end{minipage}
    \begin{minipage}{0.4\textwidth}
        \centering
        \includegraphics[width=0.66\linewidth]{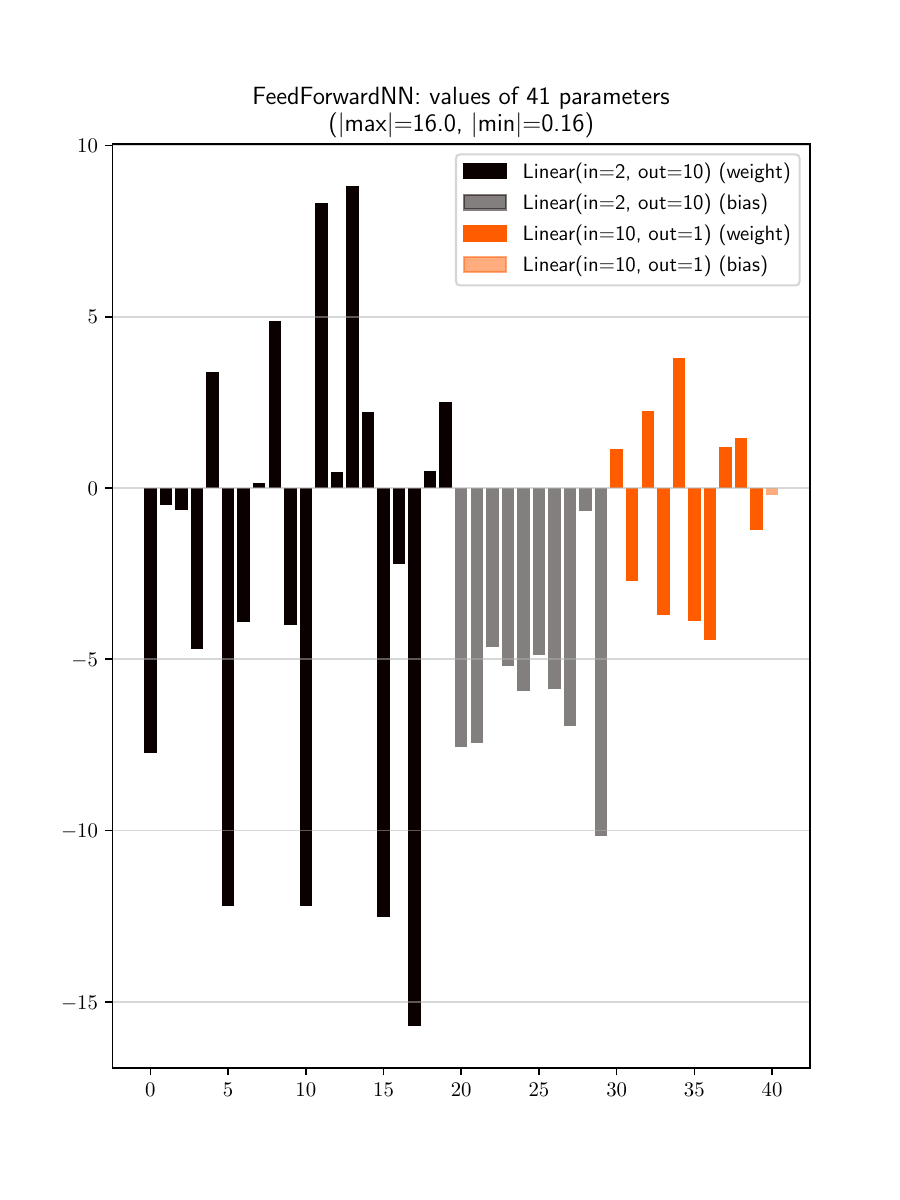}
        \captionsetup{width=0.95\linewidth}
        \caption{\small The parameters $\phi$ found by PIO for the Moons problem. All biases are negative, which may imply underutilisation.}
        \label{fig:res-weights-test}
    \end{minipage}
    
\end{adjustwidth}
\end{figure}

\begin{figure}[H]
\begin{adjustwidth}{-2cm}{-2cm}  
   \centering
   \includegraphics[width=0.3\linewidth]{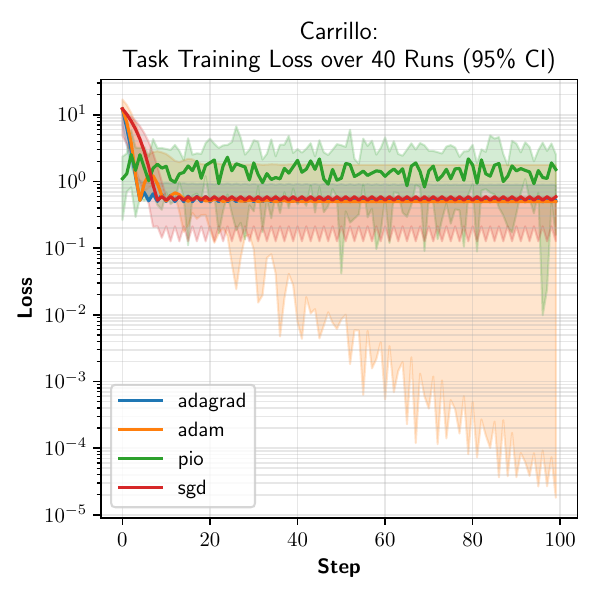}
   \includegraphics[width=0.3\linewidth]{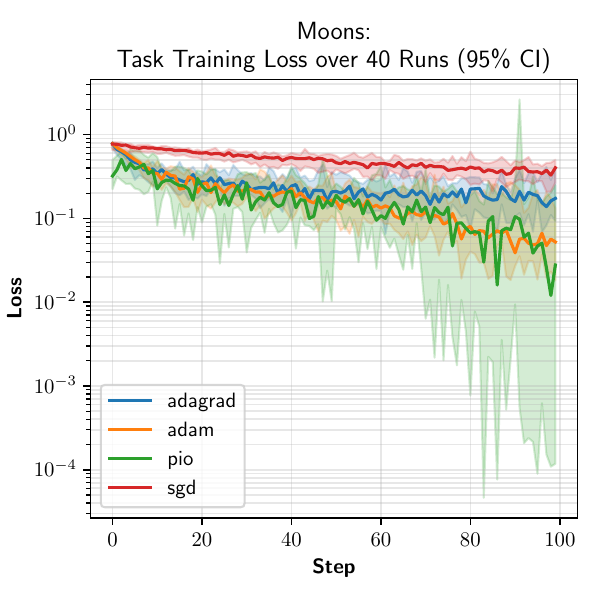} 
   \includegraphics[width=0.3\linewidth]{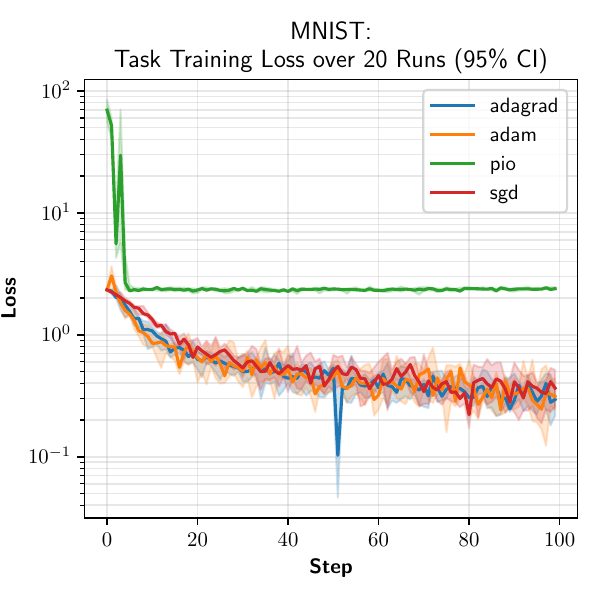} 
   \captionsetup{width=0.7\linewidth}
   \caption{\small Training loss $V(x)$ for various optimisers, tuned with a \textbf{Narrow $\times$32} hyperparameter sweep of 32 runs. Confidence intervals are computed by re-running the best-found configuration across multiple seeds.}
   \label{fig:res-losses-narrow}

\end{adjustwidth}
\end{figure}

Figure \ref{fig:res-losses-narrow} demonstrates the training loss for the best-found configurations, averaged over a number of runs, reduced for MNIST due to its computational cost.

\subsubsection{Wide $\times$64 sweep}

\begin{figure}[H]
\begin{adjustwidth}{-2cm}{-2cm}  
   \centering
   \includegraphics[width=0.3\linewidth]{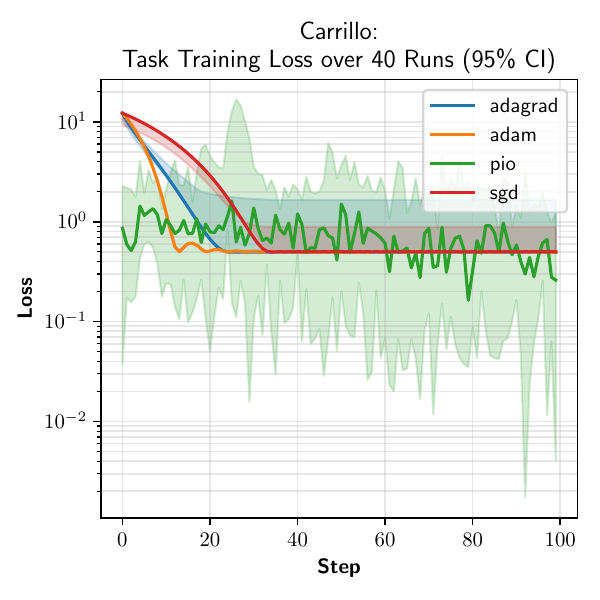}
   \includegraphics[width=0.3\linewidth]{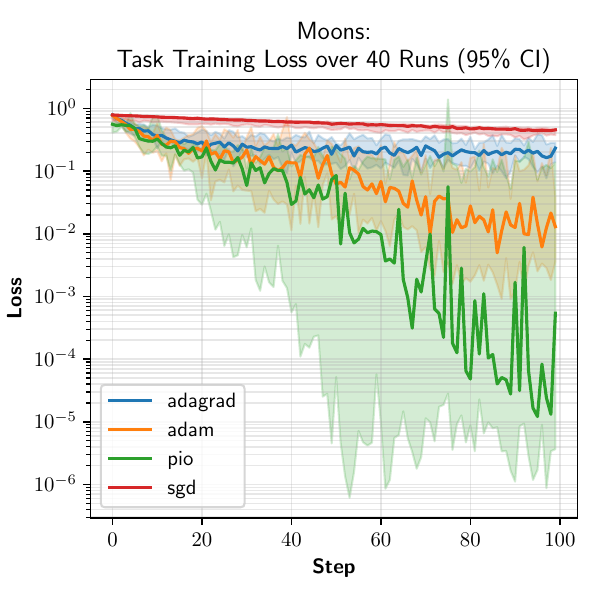} 
   \includegraphics[width=0.3\linewidth]{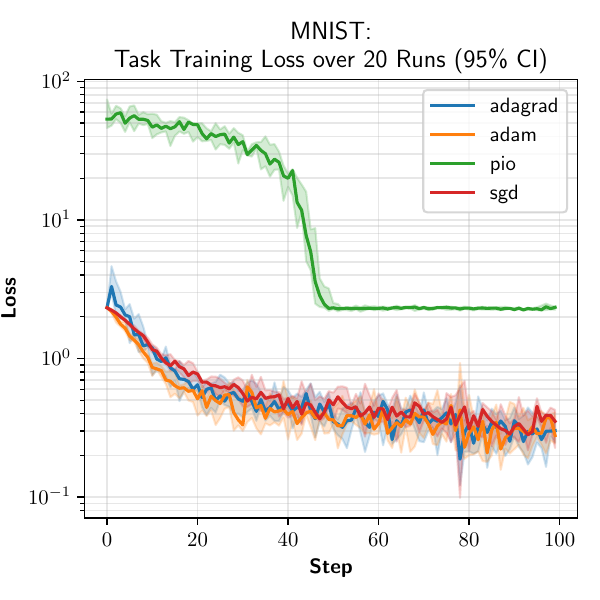} 
   \captionsetup{width=0.7\linewidth}
   \caption{\small Training loss $V(x)$ for various optimisers, tuned with a \textbf{Wide $\times$64} hyperparameter sweep of 64 runs, at learning three different task environments. Confidence intervals are computed by re-running the best-found configuration across multiple seeds. Notably, PIO still fails to learn the MNIST task.}
   \label{fig:res-losses-wide}
\end{adjustwidth}
\end{figure}

\end{document}